\documentclass[10pt, a4paper]{article}
\usepackage[]{lrec-coling2024} % this is the new style
\usepackage{natbib}
\usepackage{multibib}
\makeatletter
\usepackage{latexsym, graphicx, amsmath, float, subfig, tabularx}
\usepackage{threeparttable} % to use table notes
\usepackage{booktabs} % To thicken table lines
\usepackage{multirow}
\usepackage{multicol}
\usepackage{subcaption}

\makeatother

\usepackage{graphicx}
\usepackage{tabularx}
\usepackage{soul}
\usepackage{titlesec}
\usepackage{xcolor}
\usepackage{xstring}

\usepackage{color}

\title{Effective Distillation of Table-based Reasoning Ability from LLMs}

\name{Bohao Yang\textsuperscript{1}, Chen Tang\textsuperscript{1,2}, Kun Zhao\textsuperscript{3}, Chenghao Xiao\textsuperscript{4}, Chenghua Lin\textsuperscript{1}} 

\address{\textsuperscript{1}Department of Computer Science, The University of Manchester, UK\\
\textsuperscript{2}Department of Computer Science, The University of Surrey, UK \\
\textsuperscript{3}Department of Electrical and Computer Engineering, University of Pittsburgh, US\\
\textsuperscript{4}Department of Computer Science, The University of Durham, UK \\
         j98519by@student.manchester.ac.uk, chen.tang@surrey.ac.uk, \\
         kun.zhao@pitt.edu,
         chenghao.xiao@durham.ac.uk \\
         chenghua.lin@manchester.ac.uk\\}

\abstract{
Large Language Models (LLMs) have demonstrated remarkable performance across a wide range of natural language processing tasks. However, their enormous parameter size and extremely high requirements for compute power pose challenges for their practical deployment.
Recent research has revealed that specific capabilities of LLMs, such as numerical reasoning, can be transferred to smaller models through distillation.  Some studies explore the potential of leveraging LLMs to perform table-based reasoning. 
However, there has been no prior work focusing on table reasoning skills in smaller models specifically tailored for scientific table-to-text generation tasks.
In this paper, we propose a novel table-based reasoning distillation approach, with the aim of distilling LLMs into tailored smaller models. 
Our experimental results have shown that a 220 million parameter model (Flan-T5-base) fine-tuned using distilled data, not only achieves a significant improvement compared to traditionally fine-tuned baselines, but also surpasses specific LLMs on a scientific table-to-text generation dataset. Our code is available at \textbf{\url{https://github.com/Bernard-Yang/DistillTableCoT}}.
\\ \newline  }

\begin{document}

\maketitleabstract

% ===========================================
\vspace{2mm}
\section{Introduction}
\vspace{2mm}
% %table

Tables, as a ubiquitous and pivotal means of knowledge storage, have been receiving increasing attention in contemporary research. Tabular data, when combined with textual data, provides a valuable and complementary source of information. The intersection of tabular and textual information constitutes a well-established problem within the domain of Natural Language Processing (NLP), with impacts spanning a diverse spectrum of downstream tasks, including table question answering~\citep{Pasupat2015CompositionalSP, Cho2019ExplanatoryAA, nan_fetaqa_2022}, and table fact checking~\citep{chen_logic2text_2020, Gupta2020INFOTABSIO, Aly2021FEVEROUSFE, lu_scitab_2023}.

Conventional approaches to table-based reasoning~\citep{Pasupat2015CompositionalSP, Zhong2017Seq2SQLGS, Yu2018SpiderAL} have predominantly relied on the synthesis of executable languages such as SQL or SPARQL to facilitate information retrieval from tables. However, these symbolic languages often entail rigid assumptions regarding table structures, rendering them incapable of capturing the semantics embedded in textual segments within the table. A holistic comprehension of web tables necessitates the understanding of structured reasoning alongside textual reasoning. To this end, the emergence of table-based pre-trained models~\citep{herzig_tapas_2020, liu_towards_2021, Jiang2022OmniTabPW, Cai2022STARSG} has underscored the efficacy of pre-training models on both textual and tabular data for augmenting reasoning capabilities. This improvement stems from the extensive knowledge obtained from the large-scale crawling or synthesising of tabular and textual data.

In recent years, the advent of Large Language Models (LLMs)~\citep{Brown2020LanguageMA, Chowdhery2022PaLMSL, Touvron2023LLaMAOA} has revolutionised the landscape of NLP, ushering in a new era marked by their remarkable performance demonstrated across a multitude of controllable text generation tasks~\citep{tang-etal-2022-etrica,yang2023improving, zhao-etal-2023-evaluating, tang-etal-2023-enhancing}. Large Language Models (LLMs) implicitly capture the intricate interrelationships among tokens within input sequences, enabling them to adeptly comprehend the heterogeneous features present, regardless of their structural format, such as graph representations, tabular data, or sequential patterns\cite{huang-etal-2022-improving,goldsack-etal-2023-enhancing,tang-etal-2023-improving}. These models leverage vast corpora of textual data and undergo extensive pre-training, exhibiting an exceptional capacity to tackle intricate mathematical and commonsense reasoning tasks, often within the context of few-shot and zero-shot learning scenarios~\citep{Wei2022ChainOT, Wang2022SelfConsistencyIC, drozdov_inducing_2022,loakman-etal-2023-twistlist,zhou_least--most_2023}.

Drawing inspiration from these groundbreaking developments, a range of studies~\citep{chen_large_2023, ye_large_2023, cheng_binding_2023, Gemmell2023GenerateTA, lu_scitab_2023} have emerged to highlight the competitive performance of LLMs in comparison to state-of-the-art fine-tuned models in the domain of table reasoning tasks (e.g., table question answering and table fact-checking). For instance, \citet{zhao_large_2023} delved into the potential of employing LLMs augmented with Chain-of-Thought (CoT) techniques in the LogicNLG dataset~\cite{chen_logic2text_2020} for table-to-text generation tasks. Despite significant advancements, prior research has not focused on the challenging domain of more complex reasoning-aware scientific table-to-text generation task using LLMs. Moreover, the substantial parameter count and demanding computational requirements present obstacles to their feasible implementation.
Therefore, distilling LLMs' intrinsic table-based reasoning capabilities into more lightweight alternatives is  a more efficient and resource-friendly approach.

% ----------- fig -----------
\begin{figure}[t]
\centering
\includegraphics[width=0.99\linewidth]{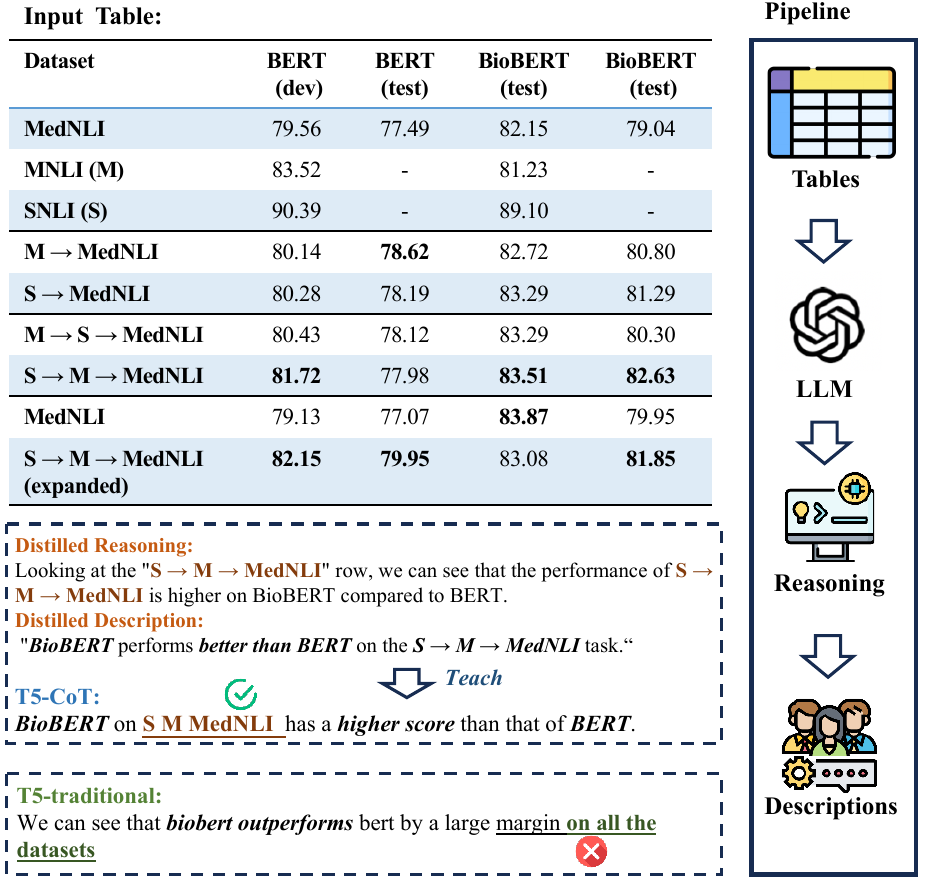}
\caption{The overview of the distillation pipeline and example data. The pipeline includes using LLMs to generate table-based reasoning and descriptions given the input table.}
\label{fig:intro}
\end{figure}
% ----------- end of fig -----------

In this paper, we investigate the capabilities of LLMs in the task of  reasoning-aware scientific table-to-text generation, and propose a two-step distillation approach to transfer the table-based reasoning ability of LLMs into smaller models. The nature of the complex scientific table-to-text generation task requires the LLMs to comprehensively grasp the provided tables and engage in arithmetic reasoning encompassing both tabular and textual data, rather than merely converting table contents into superficial descriptions. 
Our distillation pipeline is shown in Figure~\ref{fig:intro}, which includes using LLMs to generate table-based reasoning content and descriptions given the input table.
We conduct our experiments on the SciGen dataset~\cite{moosavi2021scigen}, the first scientific table-to-text dataset and is more challenging than other standard table-to-text benchmarks, such as~\citet{wiseman_challenges_2017}, \citet{, parikh_totto_2020}, and \citet{chen_logical_2020}, as it contains more numerical reasoning. 
We also provide an example in Figure~\ref{fig:intro}, in which the description generated by T5-CoT is better than that of T5-traditional, as T5-CoT is fine-tune with the reasoning and descriptions distilled from LLMs. This is because the example reasoning describes the ``\textbf{S $\rightarrow$ M $\rightarrow$ Med}'' row, which enables the model to focus on that specific row of the table in further fine-tuning the student models.

% The descriptions require the model to fully understand the given tables, and perform arithmetic reasoning over both the tabular and textual data, rather than simply converting the surface representation of the table contents. 
% We propose a two stage distillation framework containing data generation and fine-tuning stages.In the data generation stage, we utilise LLMs to generate table-based reasoning and consistency statements based on the input table, employing a one-shot CoT methodology. Subsequently, in the fine-tuning phase, we employ the distilled CoT data generated by LLMs to imbue smaller models with table reasoning proficiency.
% Our experimental results underscore that fine-tuning smaller models with table-based reasoning data distilled from LLMs leads to significant performance enhancements compared to baseline models in the context of scientific table-to-text generation tasks. 
% Distillation empowers models with as few as 220 milion parameters to outperform larger student models and even surpass the 175 billion-parameter teacher model in certain metrics.
% \textcolor{red}{[CL: I think it will be useful to summarise the paper contribution and novelty of the approch here.]}
Our contributions can be summarised as follows:
\begin{itemize}
\item We explore the potential of tackling the task of reasoning-aware scientific table-to-text generation using LLMs.
\item We propose a two-stage distillation framework containing data generation and fine-tuning stages.
In the data generation stage, we utilise LLMs to generate table-based reasoning and factually consistent statements, which could describe the table correctly based on the input table, employing a one-shot Chain-of-Thought (CoT) methodology. 
Subsequently, in the fine-tuning phase, we employ the distilled CoT data generated by LLMs to imbue smaller models with table reasoning proficiency.
% Experimental results shown fine-tuning smaller model with table-based reasoning data distilled from LLMs can achieve significant performance compare with baselines on scientific table-to-text generation task. 
% Distilling enables models as small as 0.22B to outperform larger student models, and even the 175B teacher model in certain metrics. 
\item We present a range of experimental results that underscore that fine-tuning smaller models with table-based reasoning data distilled from LLMs leads to significant performance enhancements compared to baseline models in the context of scientific table-to-text generation tasks. 
\item We demonstrate that, distillation empowers student models with as few as 220 million parameters (e.g., only 0.1\% the size of teacher model) to outperform the 175 billion-parameter teacher model in certain metrics.
% The results prove the effectiveness of AMR in improving dialogue evaluation.
\end{itemize}

\vspace{2mm}
\section{Related Work}
\vspace{2mm}

%\noindent\textbf{Table-based Reasoning.}~
\subsection{Table-based Reasoning}
Table-based reasoning tasks require the ability to reason over both natural language and structured tables.
Traditional table-based reasoning involves employing semantic parsing to execute commands on tables, with benchmarks including WikiTableQuestions~\cite{Pasupat2015CompositionalSP}, WikiSQL~\cite{Zhong2017Seq2SQLGS}, and Spider~\cite{ Yu2018SpiderAL}.
These models are designed to produce SQL for interacting with tables. However, these languages impose strict criteria on tables and make it so that these methods cannot understand the semantics of text segments.
% Therefore, a thorough understanding of language within the table is vital to improve overall performance. 
Some works proposed to learn joint representations by pre-training on table and text data~\citep{herzig_tapas_2020, liu_towards_2021, Zhao2022ReasTAPIT}. Through pre-training the model on extensive synthetic data, they are able to achieve desirable performance on table related tasks. 
Recent works~\cite{chen_large_2023, ye_large_2023, nan2023enhancing} have shown the ability of LLMs in table reasoning tasks through in-context learning. \citet{lu_scitab_2023} use LLMs to perform reasoning in the task of scientific table fact-checking. This task requires compositional reasoning using scientific tables as evidence.
BINDER~\citep{cheng_binding_2023} uses Codex to synthesise SQL queries to execute logical forms against tables in a question answering task.
% One big difference is that BINDER (Cheng et al., 2022) involves logical form execution, if the execution fails, BINDER will fall back to using language models to answer the question, which is more similar to ours.

%\noindent\textbf{Chain-of-thought Reasoning.}
\subsection{Chain-of-thought Reasoning}
Chain of thought (CoT) prompting encourages LLMs to break down a reasoning task into a series of intermediate steps, therefore enhancing reasoning abilities across various tasks~\citep{Wei2022ChainOT,Shao-etal-2024-metaphor}. With a few CoT reasoning examples, LLMs can achieve state-of-the-art performance on complex arithmetic reasoning tasks.
Self-consistency CoT~\citep{wang_self-consistency_2023} involves sampling multiple CoTs and selecting the most consistent one by beam searching. \citet{Kojima2022LargeLM} propose zero-shot CoT by first generating CoT templates and producing the final answer with LLMs in a zero-shot setting.

%\noindent\textbf{Knowledge Distillation.}~~
\subsection{Knowledge Distillation}
Distillation has demonstrated its effectiveness in transferring valuable capabilities from a larger model to a smaller one~\citep{Hinton2015DistillingTK, Sanh2019DistilBERTAD, Zeng2022GLM130BAO}.
% (Buciluaˇ et al., 2006; Hinton et al., 2015; Sanh et al., 2019; Zeng et al., 2022}. 
% The majority of distillation methods employ intermediate features or predictions from the teacher model to enhance the performance of student networks. (Wang and Yoon, 2021; Liu et al., 2021).
Recent works have shown that synthetic data generated by the teacher model can effectively transfer the specialised abilities, such as numerical reasoning, to the student model.
\citet{Chung2022ScalingIL} use manually generated CoT data to fine-tune a FLAN-based version of PaLM \cite{Chowdhery2022PaLMSL}.
\citet{fu_specializing_2023} employ enriched chain-of-thought data to specialise a smaller model. \citet{ho_large_2023} proposes diverse CoT approach by sampling different reasoning outputs from a large model to then fine-tune a smaller model.
% which endows small models with certain reasoning capabilities. 
% ~\citep{huang}
\citet{magister_teaching_2023} use a two-step pipeline for transferring the reasoning capabilities of large models to smaller models.
\citet{Hsieh2023DistillingSO} extract rationales from LLMs and integrated such data in the smaller model instruction tuning framework.
\citet{zhu_pad_2023} use LLMs to distill the programs, injecting reasoning ability into small models.
% Subsequent related work is also inspired by this concept. Hsieh et al. (2023); Chan et al. extracted rationales and integrated these data in instruction tuning framework. 
We extend the above ideas into the table-based reasoning task, specifically in the  scientific table-to-text generation domain, in which the generated CoT data leads to improved table reasoning performance.% \citet{wang_self-instruct_2023} use the generated instructions from LLMs to better instruct the model.
% Wang et al. (2020) adopted a knowledge graph complemented task to extract knowledge in LLMs. 
% Comprehensive experiments substantiate its superior training efficiency.

% ----------- fig -----------
\begin{figure*}[t]
\large
\centering
\includegraphics[width=0.99\linewidth]{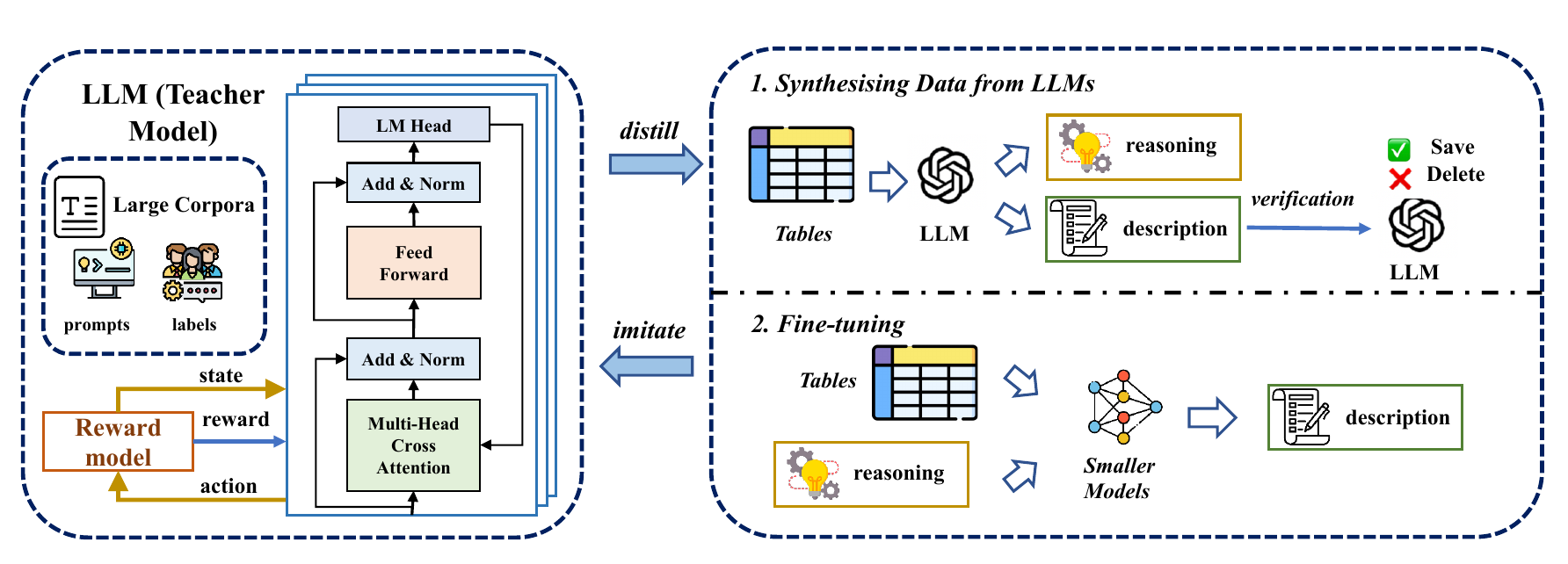}
\caption{The overview of our framework. \textbf{For synthesising data from LLMs}, we provide table examples to LLMs, and use it to generate reasonings and descriptions. Then, the generated descriptions are verified by LLMs and the false reasoning and description pairs are removed.
\textbf{For fine-tuning smaller models}, we fine-tune small models with generated reasoning and description, which inject the reasoning ability into smaller models.}
\label{fig:overview}
\end{figure*}
% ----------- end of fig -----------

\vspace{2mm}
\section{Methodology}
\vspace{2mm}

Our proposed framework is illustrated in \autoref{fig:overview}, which consists of two steps: synthesising data from LLMs and fine-tuning student models with the distilled data. The primary purpose of the first stage is to generate table-based reasoning and descriptions with LLMs given the input tables through CoT. In the second stage, the table-based reasoning ability is transferred into smaller models by fine-tuning with the distilled data from the LLMs.

\subsection{Task Definition}
We define the task as follows: The input serialised tabular data is
% is structured as a textual sequence 
denoted as $T$. In addition, the table-based reasoning data distilled from LLMs is denoted as $R = {{r_1, r_2, ..., r_n}}$, where $r_i$ is the token of reasoning. 
The primary goal of this task is to generate a description $Y = {{y_1, y_2, ..., y_m}}$, where $y_i$ is the token of the description and the model functions by simulating the conditional probability distribution $P(Y|T,R)$. The generated description should be factually consistent with the given table, and contain reasoning over the table.
% This formulation focus on generating doctor-like responses in a medical dialogue context.

\subsection{Table-based Reasoning Generation}
The data synthesis process of our proposed method is illustrated in the upper part of the right-hand side of Figure~\ref{fig:overview}, which is based on in-context learning~\cite{Brown2020LanguageMA}, an emergent ability of LLMs~\cite{Wei2022ChainOT}. Different from traditional fine-tuning, in-context learning enables the LLMs to make predictions based on the input context where only a few examples are demonstrated, without the need for parameter updating. 

We utilise a large teacher LLM, \texttt{gpt-3.5-turbo}, to generate table-based reasoning through CoT. We formulate the data generation process as follows: given a input serialised table $T$, we prompt the LLMs with the one-shot CoT demonstration example to generate a reasoning $R$ and a description $Y$ which is factually consistent with the input table. Specifically, the demonstration examples $C = {(T, R, Y)}$ is a table, reasoning, and description triplet, where the $R$ and $Y$ are hand-crafted. Finally, we can generate data as follows:
\begin{align}
R_i,Y_i = &\mathrm{LLMs}(C, T_i)
\end{align}
where we prepend the demonstrated example $C$ as the prefix to the input table $T_i$. Then the LLM will follow the instruction and learn the pattern from the example to generate corresponding reasoning $R_i$ and description $Y_i$.

\noindent\textbf{Diverse Reasoning.}
The table-to-text task enables the model to produce varied descriptions by focusing on different table regions or performing various reasoning operations, provided that the generated descriptions are factually consistent with the table~\cite{zhao_large_2023}. 
% The one-to-many issue in the open-domain dialogue generation task~\cite{gu_incorporating_2016, qiu_are_2019, Zhao2023EvaluatingOD} stated that there might be multiple reasonable response to the given conversational context. 
To maximise the reasoning ability distilled from LLMs, 
we employ the diverse reasoning approach~\cite{ho_large_2023, zhu_pad_2023, zhao_large_2023} to generate two different reasoning examples and descriptions for a given scientific table. We do not generate more reasoning-description pairs for each table because the maximal context limit of the LLMs and the average length of the tables and descriptions in the SciGen dataset is larger than in other table-to-text datasets. Specifically, the data generation process is shown as follows: given a context $C$ and table $T_i$, the LLMs are required to generate two pairs of reasoning and description.
\begin{align}
\{(R_1,Y_1), (R_2,Y_2)\}= &\mathrm{LLMs}(C, T_i)
\end{align}

\noindent\textbf{Data Filtering.}~~The synthesised table-based CoT data may contain incorrect samples due to the hallucination problem of generative models~\cite{zhu_pad_2023}. Therefore, we need to filter the wrongly generated CoT data. For filtering, we follow~\citet{madaan2023self} and employ the Self-Refine method. To be specific, when generating a new set of data $(R_i, Y_i)$ given $T_i$, we ask the LLMs to verify whether the generated description $Y_i$ is consistent with the input table $T_i$. We can filter out incorrect samples to refine our generated CoT data. The verification and filtering is crucial as the high quality training data should improve performance. Finally, we get 16,858 validated examples as the training data.

\subsection{Fine-tuning Small Models}
Once we obtain the generated table-based reasoning data, we use them to fine-tune smaller models and inject the reasoning ability into them. As for the choice of smaller models, we select T5~\cite{raffel_exploring_2019} and Flan-T5~\cite{Chung2022ScalingIL}. This is because recent works~\cite{fu_specializing_2023, zhu_pad_2023, magister_teaching_2023} have revealed that these models can attain a remarkable numerical reasoning ability when trained with CoT data in the task of complex mathematical problem solving.
We fine-tune the smaller model with the generated table-based reasoning data. 
Specifically, we concatenate the table $T$ with table-based reasoning $R$, which are split by an added special token ``<\textit{CoT}>''.
The resulting input sequence takes the
following form: ``$T$ <\textit{CoT}> $R$''. We provide an example in Figure~\ref{fig:example}.
Therefore, the description $Y$ is generated based on both the input serialised table $T$ and table-based reasoning $R$ with the following loss function:
\begin{align}
\mathcal{L}=-\frac{1}{N} \sum_{n=1}^N \log P(Y \mid T, R)
\end{align} 
where $N$ denotes the size of the training data, and $\mathcal{L}$ is the cross entropy loss.

\vspace{2mm}
\section{Experiments}
\vspace{2mm}

\subsection{Dataset}
We conduct scientific table-to-text generation on the \textbf{SciGen} dataset~\cite{moosavi_scigen_2021}. The statistics of the data are shown in Table~\ref{tab:dataset}. It consists of
three different settings: few-shot, medium and large. The train/val/test sets of medium setting are split into sizes of 13,607/3,452/1,038. The  large setting is split into 39,969/12,129/1,038. We choose the medium and large settings to conduct the experiments. This is because the few-shot setting only contains 200 examples of training data and is insufficient for fine-tuning.

\subsection{Baselines}
We follow \citet{moosavi_scigen_2021} and select T5~\cite{raffel_exploring_2019} and BART~\cite{lewis_bart_2020} as the student model baselines. For the BART baseline, we use BART-large with 0.40B parameters. For the T5 model, we use T5-base and T5-large with 0.22B, and 0.77B parameters, respectively. For the teacher models, we choose~\texttt{text-davinci-002} and \texttt{gpt-3.5-turbo} as the baseline. For the one-shot prompt setting, we follow previous works~\cite{chen_large_2023, zhao_large_2023}, which prepend one demonstration example to the input table. We compare with two variants of the teacher models, called \textit{1-shot direct} and \textit{1-shot CoT}. For the prompt formulation of 1-shot direct, we follow the setting of \citet{moosavi_scigen_2021} to linearise the table and concatenate it with the gold description as a demonstration. As for the prompt of 1-shot CoT, we prepend the input table to two hand-crafted table-based reasonings and descriptions.

\begin{table}[ht]
% \small
\centering
\begin{tabular}{lcccc}
\toprule
% \textbf{SciGen} & 
\textbf{Setting} &Text & Train & Val & Test \\
\midrule
Few-shot & 116 & 200 &100 & 1,038 \\
 Medium & 124 & 13,607 &3,452 & 1,038 \\
 Large & 133 &  39,969 &12,129 & 1,038\\ 
\bottomrule
\end{tabular}
\caption{SciGen dataset statistics. Text indicates the average length in words of descriptions.}
\label{tab:dataset}
\end{table}

% ----------- fig -----------
\begin{figure*}[]
% \small
\centering
\includegraphics[width=1.0\linewidth]{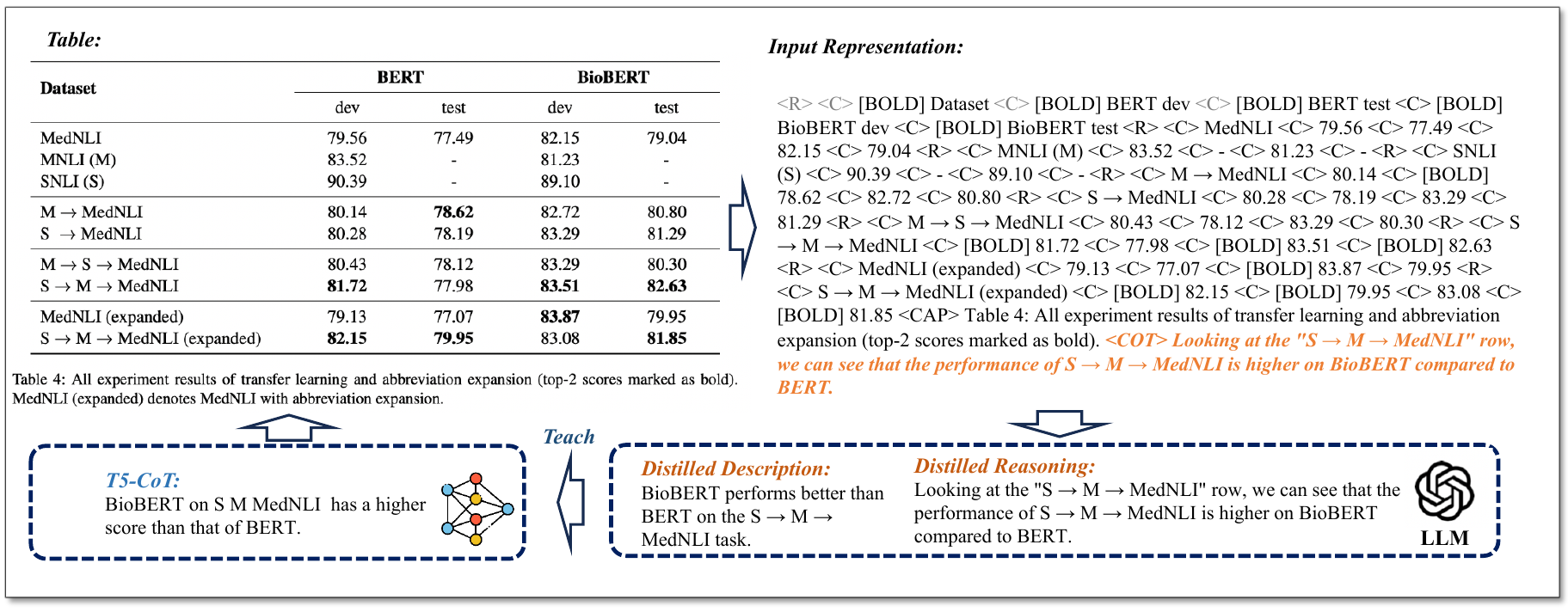}
\caption{Sample table from~\citet{nam2019surf} with its corresponding input representation. The reasoning and description are generated from LLMs for further fine-tuning smaller models.}
\label{fig:example}
\end{figure*}
% ----------- end of fig -----------

\subsection{Experimental Settings}
To use the above text-to-text generation baselines, we follow the setting in \citet{moosavi_scigen_2021} and convert tables into the text sequences. To preserve and help the model better learn the table structure, we add four special tokens to specify the beginning of rows, cells, table captions, and CoT reasoning with tokens ``<\textit{R}>'', ``<\textit{C}>'', ``<\textit{CAP}>'', ``\textit{<CoT>}'', respectively. Figure~\ref{fig:example} shows an original table from a scientific paper~\cite{nam2019surf} and its corresponding linearised input representation. The generated reasoning and description from LLMs are also provided.

% \textit{<R> <C> [BOLD] Dataset <C> [BOLD] BERT dev <C> [BOLD] BERT test <C> [BOLD] BioBERT dev <C> [BOLD] BioBERT test <R> <C> MedNLI <C> 79.56 <C> 77.49 <C> 82.15 <C> 79.04 <R> <C> MNLI (M) <C> 83.52 <C> - <C> 81.23 <C> - <R> <C> SNLI (S) <C> 90.39 <C> - <C> 89.10 <C> - <R> <C> M → MedNLI <C> 80.14 <C> [BOLD] 78.62 <C> 82.72 <C> 80.80 <R> <C> S → MedNLI <C> 80.28 <C> 78.19 <C> 83.29 <C> 81.29 <R> <C> M → S → MedNLI <C> 80.43 <C> 78.12 <C> 83.29 <C> 80.30 <R> <C> S → M → MedNLI <C> [BOLD] 81.72 <C> 77.98 <C> [BOLD] 83.51 <C> [BOLD] 82.63 <R> <C> MedNLI (expanded) <C> 79.13 <C> 77.07 <C> [BOLD] 83.87 <C> 79.95 <R> <C> S → M → MedNLI (expanded) <C> [BOLD] 82.15 <C> [BOLD] 79.95 <C> 83.08 <C> [BOLD] 81.85 <CAP> Table 4: All experiment results of transfer learning and abbreviation expansion (top-2 scores marked as bold). MedNLI (expanded) denotes MedNLI with abbreviation expansion.
% }

%%%%%%%%%%%%%%%%%%%%%%%%%%%%%%%%%%%%%%%%%%%%%%%%%%%%%%%%%%% metric %%%%%%%%%%%%%%%%%%%%%%%%%%%%%%%%%%%%%
\subsection{Automatic Evaluation Metric}
We utilise a wide range of automatic evaluation metrics from various levels to assess the performance of the model.

\noindent\textbf{Surface-level.}~~Following~\citet{moosavi_scigen_2021}, we choose \textbf{METEOR}~\cite{Banerjee2005METEORAA}, \textbf{BERTScore}~\cite{Zhang2020BERTScoreET}, and \textbf{BLEURT}~\cite{Sellam2020BLEURTLR} to measure the surface similarity of the generated statements to the gold references. 

\textbf{METEOR} aligns the output text with the reference text and computes sentence-level similarity scores based on the alignments.

\textbf{BERTScore} employs BERT embeddings, which aligns words in both the generated and reference sentences using cosine similarity. It calculates precision, recall, and F1 scores.

\textbf{BLEURT} is a learned evaluation metric based on BERT. It is first pre-trained on synthetic examples and then fine-tuned on human judgments for the task of machine translation.

However, ~\citet{moosavi_scigen_2021} stated that these metrics are not sufficient as the value range is quite low (except for BERTScore). In addition, in some cases, the incorrect description scores higher than the correct ones.

\begin{table*}
\centering
\begin{threeparttable}[b]
\resizebox{0.99\textwidth}{!}{
\begin{tabular}{l|c|cc|ccc}
\toprule[1pt]
% & \multicolumn{2}{c|}{Standard Set} & \multicolumn{2}{c}{Adversarial Set}\\ \midrule
% \textbf{Models} & \textbf{\#Params}  & \textbf{TAPAS-Acc} & \textbf{TAPEX-Acc} & \textbf{Meteor}  & \textbf{BERTScore} & \textbf{BLEURT} \\ \midrule[1pt]

\multirow{2}{*}{\textbf{Models}} & \multirow{2}{*}{\textbf{\#Params}}  & \multicolumn{2}{c|}{\textbf{Faithfulness-level}} & \multicolumn{3}{c}{\textbf{Surface-level}} \\
% \cline{3-7}
\cmidrule(lr){3-7} 
& & \textbf{TAPAS-Acc} & \textbf{TAPEX-Acc} & \textbf{Meteor}  & \textbf{BERTScore} & \textbf{BLEURT} \\ \midrule[1pt]

\multicolumn{7}{l}{\underline{\textit{Teacher Model}}}\\
\texttt{text-davinci-002 (1-shot direct)} & 175B  & 66.43 & 64.84 & 0.08 & \textbf{0.82} & -0.97\\
\texttt{gpt-3.5-turbo (1-shot direct)} & 175B  & 72.34 & 70.48 & 0.09 & \textbf{0.85} & -0.91\\
\texttt{text-davinci-002  (1-shot CoT)}  & 175B & 75.35 & 77.89 & 0.09 & 0.82 & -0.94 \\
\texttt{gpt-3.5-turbo (1-shot CoT)}  & 175B & \underline{82.53} & \underline{84.99} & 0.09 & 0.83 & -0.96 \\\midrule[1pt]

\multicolumn{7}{l}{\underline{\textit{Medium Setting}}}\\ 
BART-large & 0.40B & 57.45 & 58.41 &\textbf{ 0.23} & 0.84 & \textbf{-0.72} \\
T5-base & 0.22B & 53.27 & 52.45 & 0.15 & 0.82 & -0.89 \\
T5-large &0.77B & 56.32 & 54.78 & 0.17 & 0.83 & -0.77 \\ 
Flan-T5-base & 0.22B & 54.78 & 56.25 & 0.16 & 0.84 & -0.82 \\
Flan-T5-large &0.77B & 58.91 & 57.29 & 0.18 & 0.84 & -0.80 \\ \midrule[1pt]
% Large Setting
\multicolumn{7}{l}{\underline{\textit{Large Setting}}}\\ 
BART-large & 0.40B & 59.69 & 61.38 & 0.15 & 0.82 & -0.89 \\
T5-base & 0.22B & 55.32 & 53.76 & 0.15 & 0.82 & -0.85 \\
T5-large &0.77B & 58.21 & 56.32 & 0.18 & 0.83 & -0.79 \\
Flan-T5-base & 0.22B & 56.41 & 55.37 & 0.16 & 0.82 & -0.86 \\
Flan-T5-large &0.77B & 59.81 & 58.34 & 0.17 & 0.83 & -0.83 \\ \midrule[1pt]

\multicolumn{7}{l}{\underline{\textit{CoT fine tuning}}}\\ 
T5-base-CoT & 0.22B & 78.16 & 82.30 & 0.08 & 0.83 & -0.89 \\
T5-large-CoT &0.77B & \textbf{80.62} & 81.97 & 0.07 & 0.82 & -0.89 \\ 
Flan-T5-base-CoT & 0.22B & 78.72 & \textbf{82.75} & 0.08 & 0.82 & -0.89 \\
Flan-T5-large-CoT & 0.77B & 79.05 & 82.53 & 0.06 & 0.83 & -0.89 \\ 
\bottomrule[1pt]
\end{tabular}
}
\end{threeparttable}
\caption{Performance on the SciGen test set. Medium and large settings denote the setting of the datasets used for training. For the teacher model, \textit{direct} refers to direct prompt without CoT. \textit{CoT fine tuning} refers to fine-tuning smaller models with generated CoT data from teacher models.}

\label{tab:exp}
\end{table*}

\noindent\textbf{Faithfulness-level.}~~Recent works~\cite{moosavi_scigen_2021, liu_plog_2022} have pointed out that the above surface-level metrics cannot measure the factual correctness of the generated descriptions given the corresponding tables.
The SciGen task requires the model to generate statements which contain numerical reasoning over table values. In addition, the generated statements might cover a different table region from the gold reference. Therefore, we add two faithfulness-level metrics (to assess whether the generated sentence is grounded in the input table), \textbf{TAPAS-Acc} and\textbf{ TAPEX-Acc}~\cite{liu_plog_2022} to evaluate the factual consistency and fidelity, which have been widely used for table-to-text evaluation.
% ~\cite{liu_plog_2022, zhao_large_2023}.

\textbf{TAPAS-Acc} fine-tunes TAPAS~\cite{herzig_tapas_2020} on the TabFact dataset~\cite{chen_tabfact_2020} and achieves 81\% test accuracy.

\textbf{TAPEX-Acc} use TAPEX~\cite{liu_tapex_2022} which is fine-tuned on the TabFact dataset and achieves 84\% test accuracy.
Previous works~\cite{liu_plog_2022, zhao_large_2023} stated that TAPAS-Acc is overly positive about the
predictions, while TAPEX-Acc is more reliable for the
evaluation of the faithfulness of generated sentences.
Both above reference-free metrics score the generated descriptions as 0 for refuted and 1 for entailed given the corresponding tables.
% To evaluate the model's performance on adversarial negative examples, we created an adversarial set with 200 context-response pairs from the test set of DailyDialog++ dataset set, where the responses are adversarial negative.

% Each sampled pair was rated by three human evaluators based on two aspects: \textbf{Appropriateness}, which measures the degree to which the output is suitable within the given context, and \textbf{Coherence}, which measures the logical and meaningful presentation of the response's content. The ratings were provided on a 1-5 Likert scale, with higher scores indicating better quality. We calculated the average scores for both aspects across all annotators for each context-response pair to obtain the final human annotation score.

\begin{table*}[ht]
% \small
\centering
\resizebox{0.99\textwidth}{!}{
\begin{tabular}{lcc|lcc}
\toprule
% \textbf{SciGen} & 
\textbf{Default} &\textbf{TAPAS-Acc} &\textbf{ TAPEX-Acc} & \textbf{CoT (Ours)} & \textbf{TAPAS-Acc} & \textbf{TAPEX-Acc} \\
\midrule
T5-base & 55.32 & 53.76 & T5-base & 78.16 & 82.30\\
 Flan-T5-base & 56.41 & 55.37 & Flan-T5-base & 78.72 & 82.75\\
 \midrule
T5-large & 58.21 & 56.32 & T5-large & 80.62 & 81.97\\
 Flan-T5-large & 59.81 & 58.34 & Flan-T5-large & 79.05 & 82.53\\
\bottomrule
\end{tabular}}
\caption{Smaller model performance on the test set of the SciGen dataset. Models fine-tuned with CoT data generally perform better than the traditional fine-tuned ones (with a minimum of 20\% improvement).}
\label{tab:ablation}
\end{table*}

% ----------- fig -----------
\begin{figure*}[t]
% \small
\centering
\includegraphics[width=0.45\linewidth]{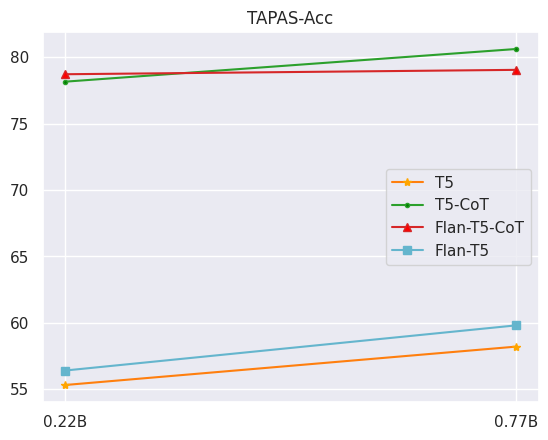}
\includegraphics[width=0.45\linewidth]{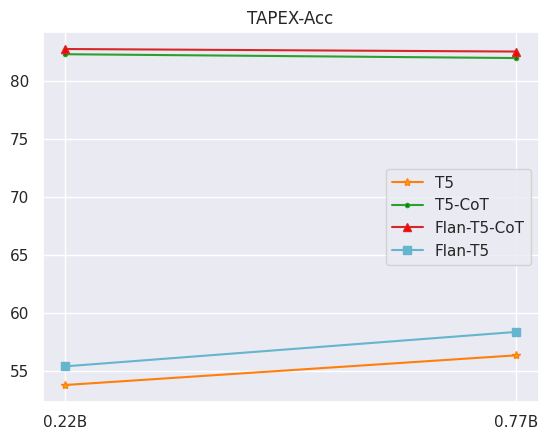}
\caption{Ablation study of smaller models on the SciGen dataset. Compared with models using standard fine-tuning, T5 and Flan-T5 fine-tuned with CoT data achieve significant improvements on both TAPAS-Acc and TAPEX-Acc.}
\label{fig:ablation}
\end{figure*}
% ----------- end of fig -----------

%%%%%%%%%%%%%%%%%%%%%%%%%%%%%%%%
%%%%%%%%%%%%%%%%%%%%%%%%%%%%%%%%Experimental results%%%%%%%%%%%%%%%%%%%%%%%%%%%%%%%%%%%%%%%%%%%%%%%%%%%%%%%%%%
\vspace{2mm}
\section{Results}
\vspace{2mm}
In this section, we evaluate both the performance of teacher LLMs and the fine-tuned smaller models on the scientific table-to-text task. We conduct automatic evaluation on both Surface-level and Faithfulness-level metrics. The overall results are shown in Table~\ref{tab:exp}. The comparison of Faithfulness-level metrics between teacher models and student models the large are presented in the Figure~\ref{fig:tapasbar} and Figure~\ref{fig:tapexbar}.

\subsection{Performance of LLMs}
Our experiments include two in-context learning methods, \textit{Direct Prompt} and \textit{CoT Prompt}. We select \texttt{text-davinci-002} and \texttt{gpt-3.5-turbo} to conduct experiments on the SciGen dataset. As shown in Table~\ref{tab:exp}, on surface-level metrics, both \textit{Direct Prompt} and \textit{CoT Prompt} cannot achieve the best performance, except for \texttt{gpt-3.5-turbo (1-shot direct)} achieving the best performance on BERTScore. However, the surface-level metrics are unable to accurately measure the faithfulness and accuracy of the models' generated outputs. In terms of the faithfulness-level metrics, 
\texttt{text-davinci-002 (1-shot direct)} can achieve over 64\% accuracy and \texttt{gpt-3.5-turbo (1-shot direct)}
can achieve over 70\% accuracy on both TAPAS-Acc and TAPEX-Acc, which outperform the traditional fine-tuned baseline models (i.e. BART and T5). When combined the direct prompt with CoT reasoning, the accuracy of both \texttt{text-davinci-002 (1-shot CoT)} and \texttt{gpt-3.5-turbo (1-shot CoT)} increases by around 10\% on both metrics.
% When prompting the LLMs with CoT, there is an about 10\% accuracy improvement in faithfulness-level metric compared with the \textit{Direct Prompt}.% The aim of the model is to discriminating the positive context-response pairs from the adversarial examples. 
% We conduct the experiments on the training and testing sets during the training and testing phase, respectively.

\subsection{Performance of Fine-tuned Smaller Model}
Regarding the surface-level metrics, the smaller models, whether fine-tuned with CoT data or not, consistently exhibit a narrow range of low values, with absolute values falling within the 0-1 Likert scale range. The experimental results are consistent with the statements in SciGen's paper~\cite{moosavi_scigen_2021} that surface-level metrics are not sufficient to reflect models' abilities on this complex task.

% ----------- fig -----------
\begin{figure}[]
% \small
% \centering
\includegraphics[width=0.99\linewidth]{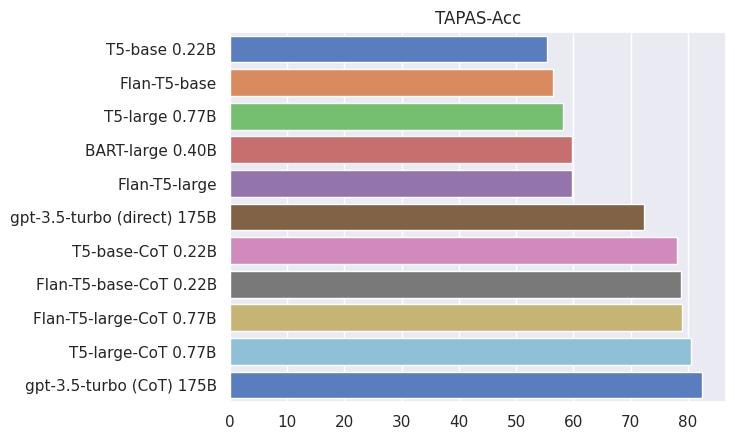}
\caption{The TAPAS-Acc of the teacher models (LLMs) and small models on the SciGen dataset. All the small models fine-tuned with CoT data can surpass LLMs with direct prompting.}
\label{fig:tapasbar}
\end{figure}
% ----------- end of fig -----------

% ----------- fig -----------
\begin{figure}[]
% \small
\centering
\includegraphics[width=0.99\linewidth]{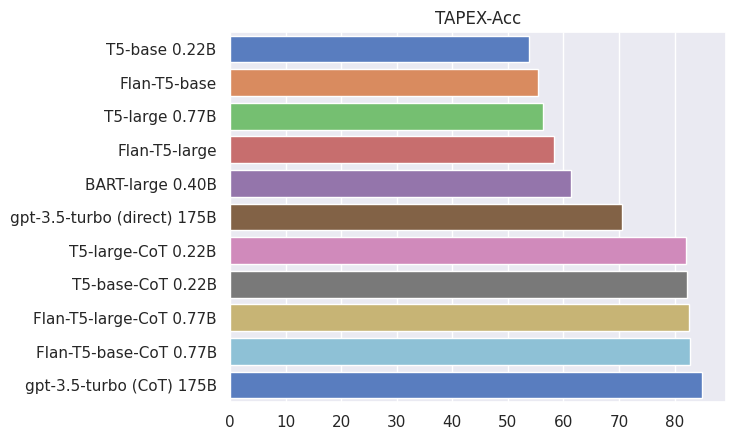}
\caption{The TAPEX-Acc of teacher models and small models on the SciGen dataset. The trend is similar to TAPAS-Acc, with the performance of small models fine-tuned with CoT data only underperforming when compared to LLMs with CoT prompting.}
\label{fig:tapexbar}
\end{figure}
% ----------- end of fig -----------

\noindent\textbf{Small models with traditional fine-tuning do not perform well on faithfulness-level metrics.}~
In terms of the smaller models fine-tuned without CoT data, BART-large fine-tuned on the medium dataset achieves the best on surface-level metrics. However, in terms of the faithfulness-level, all the BART and T5 baselines only achieve an accuracy slightly higher than random chance.
We further investigate the impact of dataset size, ranging from the \textit{Medium Setting} to the \textit{Large Setting}. Although the size of the Large Setting dataset is three times that of the Medium Setting, performance improvements are not as significant (i.e., only around 2\% increase on the faithfulness-level metrics). However, for the surface-level metrics, models that are trained with the Medium datasets achieve better overall performance, especially in METEOR and BLEURT.

\noindent\textbf{Small models fine-tuned with CoT data achieve a significant performance improvement.}~
On the other hand, the T5 and Flan-T5 models with CoT fine-tuning can achieve the best overall performance on the faithfulness-level metrics among all the small models. All the performances of CoT fine-tuning models are on par with the teacher model (i.e., \texttt{gpt-3.5-turbo (1-shot CoT)} on the faithfulness-level metrics. For instance, T5-large-CoT and Flan-T5-base-CoT achieve the highest TAPAS-Acc (80.62\%) and TAPEX-Acc (82.75\%), and only underperform the teacher model with the best performance by a margin of 2\%. These results indicate that fine-tuning with CoT data distilled from LLMs can transfer the table-based reasoning ability into smaller models.

\noindent\textbf{Larger model size does not guarantee the performance improvement when fine-tuned without CoT data.}~
Furthermore, our experiments also investigate the impact of the model size for CoT fine-tuning, ranging from the base to the large variant. While it is intuitive to expect performance improvements with larger models, the experimental results on TAPEX-Acc metric reveal that models with larger parameter counts, such as T5-large and Flan-T5-large, do not consistently outperform their smaller counterparts, T5-base and Flan-T5-base. However, regarding TAPAS-Acc, the performance improvement is consistent, with the model size increasing from base (0.22B) to large (0.77B).

\subsection{Comparison between Teacher and Student Models}
We also compare the performance on faithfulness-level metrics (TAPAS-Acc and TAPEX-Acc) of both the teacher model (LLMs) and student models in Figure~\ref{fig:tapasbar} and Figure~\ref{fig:tapexbar}. For the teacher model, \texttt{gpt-3.5-turbo (1-shot direct)} outperforms all smaller baseline models (smaller models fine-tuned without CoT data) and \texttt{text-davinci-002 (1-shot direct)}. In addition, \texttt{gpt-3.5-turbo (1-shot CoT)} achieves the best performances on both TAPAS-Acc and TAPEX-Acc metrics among both teacher and student models. As for smaller models, both T5 and Flan-T5 can only achieve around 55\% accuracy on both faithfulness-level metrics without being fine-tuned with CoT data. However, these smaller models can be injected with reasoning ability after fine-tuning with CoT data, achieving approximately 80\% accuracy on both metrics.

\begin{figure}[ht]
\small
\centering
\includegraphics[width=0.99\linewidth]{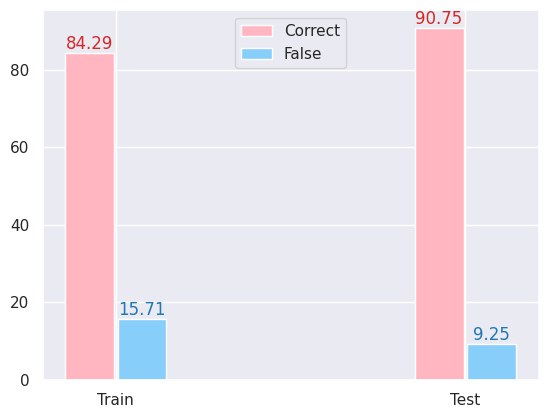}
\caption{Evaluation of generated data of train and test sets of SciGen dataset. Correct refers to the data with statements verified correctly by LLMs.}
\label{fig:data}
\end{figure}

\subsection{Ablation Study}
The ablation study of fine-tuning with CoT data are shown in both Table~\ref{tab:ablation} and Figure~\ref{fig:ablation}. For both T5 and Flan-T5 models, we can observe the significant increases after fine-tuning with with CoT data in both TAPAS-Acc and TAPEX-Acc on the SciGen table-to-text generation task. 
For TAPAS-Acc metric, T5 and Flan-T5 base (0.22B) and large (0.77B) models can only achieve over 55\% accuracy. However, when fine-tuning with table-based CoT data from LLMs, there is a significant accuracy increase (over 20\%) observed. For instance, the 55\% accuracy of T5-large with standard fine-tuning can be improved to 80\% after being fine-tuned with CoT data.
As for TAPEX-Acc metric, a similar trend can be observed, where the overall improvement in accuracy is over 25\%. For example, the most significant improvement can be observed in the T5-base model, which is from 53\% (traditional fine-tune) to 82\% (CoT fine-tune).

% When compared to standard fine-tuning, PaD presents a considerable advantage. 
% % ----------- fig -----------
% \begin{figure}[]
% \small
% \centering
% \includegraphics[width=0.88\linewidth]{TAPAS-Acc.png}
% \caption{The ablation study on TAPAS-Acc. }
% \label{fig:tapas-t5}
% \end{figure}
% % ----------- end of fig -----------

% % ----------- fig -----------
% \begin{figure}[]
% \small
% \centering
% \includegraphics[width=0.88\linewidth]{TAPEX-Acc.png}
% \caption{The ablation study on TAPAS-Acc. }
% \label{fig:tapas-t5}
% \end{figure}
% % ----------- end of fig -----------
% ----------- fig -----------

% ----------- end of fig -----------

\subsection{Generated Data Analysis}
The LLMs we used in this paper contributed towards the synthesis of high-quality table-based CoT data. However, during the generation process, there are certain falsely generated data due to the hallucinatory nature of LLMs.
Therefore, we conduct a comprehensive analysis of the samples generated by LLMs. The evaluation results are shown in Figure~\ref{fig:data},~\texttt{gpt-3.5-turbo} achieves an accuracy of 85\% on the training set, where the generated descriptions are verified as correct. As for the test set of SciGen, the accuracy is over 90\%, and with less than 10\% of the samples regarded as incorrect. Regarding the table-to-text generation task, both the generated reasoning and descriptions reveal high-quality coherence and consistency given the input table.

\vspace{2mm}
\section{Conclusion}
\vspace{2mm}
In this paper, we introduce a two-stage distillation framework that distills table-based CoT data from LLMs. Our experiments illustrate that this method is able to effectively transfer table reasoning abilities to smaller models in the scientific table-to-text generation task. The performance improvement can even outperform certain teacher LLMs (e.g.,~\texttt{gpt-3.5-turbo}). Our proposed method achieves comprehensive superiority in this specific task while requiring less data and smaller models. 
% Further analysis reveals that PaD possesses higher training efficiency and is better suited for smaller models.
\vspace{2mm}
\section*{Bibliographical References}
% \section{Appendix: How to Produce the \texttt{.pdf}}
% \label{sec:append-how-prod}
% In order to generate a PDF file out of the LaTeX file herein, when citing language resources, the following steps need to be performed:
\bibliographystyle{lrec-coling2024-natbib}
\bibliography{zotero, lrec-coling2024-example}

\end{document}